\documentclass[conference]{IEEEtran}
\IEEEoverridecommandlockouts
\usepackage{cite}
\usepackage{amsmath,amssymb,amsfonts}
\usepackage{algorithmic}
\usepackage{graphicx}
\usepackage{textcomp}
\usepackage{xcolor}
\usepackage{array}
\usepackage{geometry}

\def\BibTeX{{\rm B\kern-.05em{\sc i\kern-.025em b}\kern-.08em
    T\kern-.1667em\lower.7ex\hbox{E}\kern-.125emX}}
\begin{document}
\newgeometry{top=25mm,
    left=19mm,
    right=19mm,
    bottom=19mm,} 
    

\title{Study and Survey on Gesture Recognition Systems\\
}

\author{\IEEEauthorblockN{Kshitij Deshpande}
\IEEEauthorblockA{\textit{Information Technology} \\
\textit{Pune Institute of Computer Technology}\\
Pune, India \\
kshitij.deshpande7@gmail.com}
\and
\IEEEauthorblockN{Varad Mashalkar}
\IEEEauthorblockA{\textit{Information Technology} \\
\textit{Pune Institute of Computer Technology}\\
Pune, India \\
varadmash2201@gmail.com}
\and
\IEEEauthorblockN{Kaustubh Mhaisekar}
\IEEEauthorblockA{\textit{Information Technology} \\
\textit{Pune Institute of Computer Technology}\\
Pune, India \\
kaustubh.m0803@gmail.com}
\and 
\IEEEauthorblockN{Amaan Naikwadi}
\IEEEauthorblockA{\textit{Information Technology} \\
\textit{Pune Institute of Computer Technology}\\
Pune, India \\
amaannaikwadi@gmail.com}
\and
\IEEEauthorblockN{Dr. Archana Ghotkar}
\IEEEauthorblockA{\textit{Information Technology} \\
\textit{Pune Institute of Computer Technology}\\
Pune, India \\\
aaghotkar@pict.edu}
\and
}
\maketitle

\begin{abstract}

In recent years, there has been a considerable amount of research in the Gesture Recognition domain, mainly owing to the technological advancements in Computer Vision. Various new applications have been conceptualised and developed in this field. This paper discusses the implementation of gesture recognition systems in multiple sectors such as gaming, healthcare, home appliances, industrial robots, and virtual reality. Different methodologies for capturing gestures are compared and contrasted throughout this survey. Various data sources and data acquisition techniques have been discussed. The role of gestures in sign language has been studied and existing approaches have been reviewed.
Common challenges faced while building gesture recognition systems have also been explored.
\end{abstract}

\begin{IEEEkeywords}
gestures, gesture recognition, computer vision, sign language
\end{IEEEkeywords}

\section{Introduction}
Gestures are nonverbal movements and signals made by the body, such as hand movements, facial expressions, and body posture. They are an essential part of human communication and can convey a wide range of meanings and emotions without the use of words. Gestures can be used in HCI applications as they can provide an intuitive way for users to interact with technology. This can reduce the learning curve for new users and make the interaction more accessible to a broader range of users. Gestures can also be a more efficient way of completing actions and interacting with the system. They can be an effective way to provide an accessible interaction for users with physical disabilities
or impairments. For example, using gestures for navigation can be easier for users with limited mobility than using a mouse or keyboard. Gestures facilitate communication with people who do not share a common language. Also for people who are disabled and cannot communicate verbally, sign language is used.
\newline
Sign language is a visual language that makes use of a combination of hand gestures, facial expressions, and body movements to communicate with others who are deaf or hard of hearing. Hand gestures, also known as signs, are the most important part of sign language. Each sign corresponds to a word or concept, and they are combined to form sentences and express complex ideas. For example, the sign for “allergy” in the Indian Sign Language (ISL) involves using one hand to scratch the other to convey the word “allergy”. Body posture and movement are also important in sign language, for example, to convey the sign “hospital” in Indian Sign Language one needs to turn sideways and draw a “+” symbol on the side of their shoulder. Facial expressions are also used in sign language, as
they can convey emotional nuance and tone.
\newline
In this study, we describe the architecture and operation of a general gesture recognition system. We will look at the different methodologies that can be used for detecting and interpreting gestures by a system as well as various applications of gesture detection. A comparative analysis of various tools and techniques for recognising gestures, along with the comparison of various data sources for the same is also conducted. We look at various benchmark datasets in this domain and review multiple technologies and algorithms that have been implemented in gesture recognition systems, along with the challenges faced while doing so. We understand need of using gestures for communicating
using sign language and its applications as well.

\section{Gesture Recognition Systems and Datasets}
 Gesture recognition systems are an increasingly important area of research that enables natural and intuitive interaction between humans and machines. These systems rely on high-quality datasets to train and improve their accuracy, making dataset creation and curation a crucial aspect of gesture recognition research. In this review, we will explore some of the most notable gesture recognition systems and datasets in the field.

 \subsection{Algorithms}
 Eng-Jon, et al [1] propose to model the spatio-temporal signatures using Sequential Interval Patterns (SIP), which is based on the Greek Sign Language. A novel multi-class classifier is then used to organise these SIP into a hierarchical 
\restoregeometry





\begin{table*} [htbp]
\normalsize
\begin{center}
\caption{Comparison of Existing Sign Language Datasets}
\begin{tabular}{|p{1cm}|p{3cm}|p{2cm}|p{2cm}|p{2cm}|p{2cm}|p{2cm}|}
 \hline
  Sr no. & Name & Country & Classes & Subjects & Samples & Language Level\\ [0.5ex] 
 \hline
 1 & SIGNUM & Germany & 450 & 25 & 33210 & Sentence  \\
 \hline
 2 & GSL 20 & Greek & 20 & 6 & ~840 & Word  \\
 \hline
 3 & Boston ASL LVD & USA & 3300+ & 6 & 9800 & Word  \\
 \hline
 4 & PSL Kinect 30 & Poland & 30 & 1 & 300 & Word  \\
 \hline 
 5 & LSA64 & Argentina & 64 & 10 & 3200 & Word  \\
 \hline
 6 & MSR Gesture 3D & USA & 12 & 10 & 336 & Word  \\
 \hline
 7 & DEVISIGN-L & China & 2000 & 8 & 24000 & Word  \\
 \hline
 8 & LSFB-CONT & Belgium & 6883 & 100 & 85000+ & Word / Sentence  \\
 \hline
 9 & LSFB-ISOL & Belgium & 400 & 85 & 50000+ & Word  \\
 \hline
 10 & WLASL & EEUU & 2000 & 119 & 21083 & Word  \\
 \hline
\end{tabular}
\end{center}
\end{table*}
form which makes it possible to take advantage of any sub-sequence sharing that may occur between various SIPs of various classes. Dilsizian, Mark, et al. [2] put forward a new framework based on the American Sign Language that offers a new tracking technique that is less reliant on laboratory settings and can handle variations in background and skin regions (like the face, forearms, or other hands) , identifies 3D hand configurations that are significant for ASL linguistically and  includes statistical data that reflects linguistic restrictions on sign creation. Oszust, M., and M. Wysocki. [3] compares the  strategy using models of subunits to a complete word model approach using the closest neighbour classifier which uses dynamic time warping (DTW) based on the Polish Sign Language. Kinect's skeletal pictures were used for one set of features and simplified descriptions of the hands retrieved as skin-colored regions were employed for the other. The second strategy proved to be more effective.\\\\
Masood, Sarfaraz, et al. [4] implement a very straightforward approach to Argentinian sign language recognition in video sequences, which consists of temporal and spatial features. RNNs have been used for the former while CNNs have been used for the latter. An accuracy of 95.217 was achieved.
Li, Dongxu, et al. [5] suggests a strategy that involves exploiting cross-domain knowledge in subtitled sign news videos to enhance sign language recognition models based on word-level American Sign Language. A domain-invariant descriptor is learned for each isolated sign, class centroids are stored in prototypical memory, and isolated signs and news signs are aligned. This improves performance and reduces the need for labelling by enabling the classifier to concentrate on learning class-specific features and localise sign words from sentences automatically.\\\\
A general probabilistic model for categorising signs is presented by Ronchetti, Franco, et al. [6] and includes sub-classifiers based on numerous variables, including location, movement, and handshape in the Argentinian Sign Language. In all categorization steps, the model employs a bag-of-words strategy. The suggested model had a 97 percent accuracy rate when tested on a dataset of Argentine Sign Language with 64 classes of signs and 3200 samples, providing some evidence that recognition without the sequential ordering of frames to limit the number of hypotheses is possible.
\\

\subsection{Related Work}
\vspace{5pt}
Bokstaller, Jonas, and Costanza Maria Improta. [7] present a basic fundamental approach towards dynamic sign language recognition by simply training a vanilla Conv3DRNN network on dynamic signs. It consists of a LSTM layer that is fed by a 3D-CNN, performing with a best accuracy of around 77 percent. Chen, Yutong, et al. [8] demonstrate an approach, in which two different streams are used to model RGB videos and keypoint sequences. These are then used together for sign language recognition by applying techniques like bidirectional lateral connection which does an element wise addition between the two feature maps. A sign pyramid network (SPN) is defined that involves a top-down pathway and a lateral connection to generate frame level probabilites.
Hu, Hezhen, et al. [9] modelled an algorithm called SignBERT which is a self-supervised pretrainable sign language recognition system with model-aware hand prior incorporated. In the first self supervised pre training step it masks and reconstructs the hand, and captures the temporal context of the sign. In the second step the model is fine tuned to the task of sign language recognition by replacing the decoder from the intial model by a prediction head. Boh aˇcek, Maty aˇs, and Marek Hr uz. [10] presented a paper that shows a new approach for word-level sign language recognition based on a sign pose-based transformer model. The proposed method takes sign language videos as input and extracts pose information to represent the sign language signs. The extracted pose information is then fed into a transformer model that learns to recognize the corresponding sign language word. 
\\\\
Zuo, Ronglai, Fangyun Wei, and Brian Mak. [11] demonstrated an approach in which a new method for sign language recognition that utilizes natural language processing (NLP) techniques is proposed. The proposed system takes advantage of the fact that sign languages often have a natural language equivalent and leverages this to improve accuracy in recognizing signs. The system uses a pre-trained NLP model to generate possible natural language phrases for a given sign, and then compares these phrases to a database of known sign language phrases to determine the correct sign. 
\\\\
Lai, Kenneth, and Svetlana N. Yanushkevich [12] propose a method that uses the depth and skeleton information of the hand to detect and classify hand gestures. The method consists of two main stages: a pre-processing stage and a gesture recognition stage. In the pre-processing stage, the depth and skeleton data of the hand are extracted using a Kinect sensor. Then, the data is normalized and segmented to extract individual gestures. In the gesture recognition stage, the normalized gesture data is fed into a CNN to extract features, followed by an RNN to capture the temporal dynamics of the gesture. The output of the RNN is then fed into a softmax classifier for gesture recognition. Alam, Mohammad Mahmudul, Mohammad Tariqul Islam, and SM Mahbubur Rahman. [13] also demostrate a unified learning approach for simultaneous recognition of egocentric hand gestures and detection of fingertips in first-person videos. The proposed method utilizes a single deep learning network that takes raw video frames as input and produces hand gesture labels and fingertip locations as outputs. The model is trained on a large-scale egocentric hand gesture dataset and achieves state-of-the-art performance in both hand gesture recognition and fingertip detection tasks.
\\\\
Chen, Yuxiao, et al. [14] propose a method for recognizing hand gestures using dynamic graphs constructed with spatial-temporal attention. The proposed method creates a dynamic graph from extracted features where nodes represent the features and the edges represent the relation between them. A spatial-temporal attention mechanism is employed to assign weights to the nodes and edges.

Chen, Huizhou, et al. [15] demonstrate a new method for recognizing hand gestures using a combination of 3D CNNs and multi-scale attention mechanisms. The proposed network takes in multimodal data from depth and color cameras, and learns to extract features at different spatial and temporal scales. The network performs better overall when the multi-scale attention mechanism is employed to selectively emphasise key features at various scales.

\section{Features used in gesture recognition}

\subsection{Types of gestures}
\subsubsection{Static gestures}
Static gestures involve holding the hand or body in a specific pose or configuration without any movement. Some common features used in static gesture recognition include:
\begin{table}[htbp]
\normalsize
\begin{center}
\caption{Techniques for feature extraction for static signs}

\begin{tabular}{|p{2cm}|p{3cm}|p{2cm}|}
\hline
Feature & Technique & Tool \\ \hline
Hand shape & Shape analysis & OpenCV, MATLAB \\ \hline
Palm orientation & Normalized depth map & OpenCV, MATLAB \\ \hline
Hand position & Convex hull and centroid detection & OpenCV, MATLAB \\ \hline
Distance & Euclidean distance between hand and other objects & OpenCV, MATLAB \\ \hline
Texture & Local binary patterns (LBP) & OpenCV, MATLAB \\ \hline
Color & Color histogram of selected regions & OpenCV, MATLAB \\ \hline
Saliency & Saliency map analysis & OpenCV, MATLAB \\ \hline
\end{tabular}
\end{center}
\end{table}

\subsubsection{Dynamic gestures}
Dynamic gestures are gestures that involve movement of the hand or body. Some of the features used in dynamic gesture recognition are:
   
\begin{table}[htbp]
\normalsize
\caption{Techniques for feature extraction for dynamic signs}
\begin{center}

\begin{tabular}{|p{1.8cm}|p{3cm}|p{2.5cm}|}
\hline
Feature & Technique & Tool \\ \hline
Trajectory & Motion tracking & OpenCV, MATLAB \\ \hline
Velocity & Finite difference method & OpenCV, MATLAB \\ \hline
Acceleration & Finite difference method & OpenCV, MATLAB \\ \hline
Jerk & Finite difference method & OpenCV, MATLAB \\ \hline
Curvature & Curvature estimation & OpenCV, MATLAB \\ \hline
Hand shape & Shape analysis & OpenCV, MATLAB \\ \hline
Joint angles & Kinematic modeling & OpenCV, MATLAB \\ \hline
Frequency & Fourier transform & OpenCV, MATLAB \\ \hline
\end{tabular}
\end{center}
\end{table}

\subsection{Method of recognition}
There are primarily 2 ways of recognising gestures, Vision based and Sensor based. Vision-based gesture recognition systems rely on image and video processing techniques. Sensor-based gesture recognition systems typically use a combination of hardware sensors and signal processing algorithms to capture and analyze gesture data.
 
\begin{table}[htbp]
\normalsize
\begin{center}
\caption{Techniques for feature extraction in vision-based recognition}

\begin{tabular}{|p{1.8cm}|p{3cm}|p{2.5cm}|}
\hline
Feature & Technique & Tool \\ \hline
Hand and body pose & Pose estimation & OpenCV, TensorFlow \\ \hline
Hand and body movement & Motion tracking & OpenCV, MATLAB \\ \hline
Skin color and texture & Color histogram, LBP & OpenCV, MATLAB \\ \hline
Edge detection & Canny edge detection & OpenCV, MATLAB \\ \hline
Optical flow & Lucas-Kanade method & OpenCV, MATLAB \\ \hline
Depth information & Structured light, Time-of-flight camera & OpenCV, MATLAB \\ \hline
Hand landmarks & Landmark detection & OpenCV, TensorFlow \\ \hline
\end{tabular}
\end{center}
\end{table}


\begin{table}[htbp]
\normalsize
\begin{center}
\caption{Techniques for feature extraction for sensor-based recognition}

\begin{tabular}{|p{1.8cm}|p{3cm}|p{2.5cm}|}
\hline
Feature & Technique & Tool \\ \hline
Inertial measurement units & Kalman filtering, quaternion & MATLAB, Python \\ \hline
Acceleration & Signal processing & MATLAB, Python \\ \hline
Gyroscopic rotation & Signal processing & MATLAB, Python \\ \hline
Magnetic field & Signal processing & MATLAB, Python \\ \hline
Flex sensors & Resistance measurement & Arduino, Raspberry Pi \\ \hline
Pressure sensors & Resistance measurement & Arduino, Raspberry Pi \\ \hline
Infrared sensors & Reflection measurement & Arduino, Raspberry Pi \\ \hline
Ultrasonic sensors & Time-of-flight measurement & Arduino, Raspberry Pi \\ \hline
\end{tabular}
\end{center}
\end{table}

In summary, the features used in gesture recognition depend on the type of gesture and the method of recognition. Static gestures involve hand shape, orientation, and palm orientation, while dynamic gestures involve trajectory, velocity, and acceleration. Vision-based recognition involves shape analysis, motion analysis, and texture analysis, while sensor-based recognition involves IMUs, flex sensors, and pressure sensors

\section{Methodologies for capturing gestures}

There are multiple methodologies identified for capturing gestures in a computational environment. These methodologies can utilize various available tools and provide a set of features as required. The selection of the required methodology can be based on the provided features. Table VI of this paper elaborates the features and challenges associated with these methodologies along with their applications in various sectors

\begin{table*}[htbp]
\normalsize
\caption{Comparison of Data Collection Methods}
\begin{center}
\begin{tabular}{|p{3cm}|p{3.5cm}|p{3cm}|p{3.5cm}|p{3cm}|}
\hline
\textbf{Source of Data Collection} & \textbf{Challenges} & \textbf{Tools} & \textbf{Features} & \textbf{Applications} \\
\hline
Camera-based systems & - Limited field of view & - Webcams, Kinect & - High accuracy & - Human-computer interaction \\
& - Poor lighting conditions & - Leap Motion & - Can track multiple joints & - Sign language recognition \\
& - Occlusions & - Intel RealSense & - Can capture RGB and depth information & - Gaming \\
& - Privacy concerns & - OpenPose & - Can recognize body pose and movement & - Sports training \\
\hline
Wearable devices & - Limited battery life & - Smartwatches, armbands & - Continuous data collection & - Fitness tracking \\
& - Comfort and fit & - Smart gloves & - Can capture hand gestures & - Industrial control \\
& - Limited sensor range & - Myo armband & - Can recognize muscle activation & - Virtual reality \\
& - Complex data processing & - Nymi wristband & - Can authenticate user based on heart rate & - Healthcare \\
\hline
Inertial measurement units & - Calibration errors & - Accelerometers, gyroscopes & - Low power consumption & - Robotics \\
& - Drift and noise & - Magnetometers & - Small form factor & - Motion analysis \\
& - Limited accuracy & - IMU sensor fusion & - Can provide 3D orientation & - Gait analysis \\
& - Sensor placement & & & \\
\hline
Depth sensors & - Limited range & - Kinect, Intel RealSense & - Can capture 3D information & - Gaming \\
& - Poor lighting conditions & - Structure Sensor & - Can attach to mobile devices & - Robotics \\
& - Occlusions & - Zed Camera & - Can capture RGB and depth information & - Augmented reality \\
& - Complexity and cost & - Orbbec Astra & - Can track multiple users & - Human-computer interaction \\
\hline
Electromyography & - Limited sensor placement & - Surface EMG sensors & - Can capture muscle activation & - Prosthetics \\
& - Signal interference & - Intramuscular EMG sensors & - High accuracy & - Rehabilitation \\
& - Limited range & - Wireless EMG sensors & - Continuous data collection & - Sports training \\
& - Complex data processing & & & \\
\hline
\end{tabular}
\end{center}
\end{table*}

\section{Gesture based applications}

\subsection{Gaming}

Gesture based interfaces provide the users with a better interaction than tradition controls. Gestures can be used to control actions, directions and make choices. However, the gesture recognition needs to be fast and accurate to provide an absolute experience.


Vision based methods can identify gestures to reproduce the same in a gaming environment. Nintendo Wii, Sony Playstation Eye, Microsoft Kinect and Leap Motion are some popular gesture control devices used in the gaming industry.

\subsection{Healthcare}

FAce MOUSe was introduced by Nishikawa et al. [16], which uses the facial expressions of surgeons to control the position of laparoscope, thus eliminating the use of direct contact devices. Wachs et al. [17] invented Gestix, a real-time, vision-based hand gesture interpretation system that employs the user's gestures to navigate and manipulate pictures in an electronic medical record (EMR) database. Through captured video and navigation, gestures are converted to instructions based on their temporal trajectories.

\subsection{Home Appliances}

Users can lock or open doors, change camera angles, arm or disarm security systems, and receive alerts when someone is detected all with the use of simple hand gestures. With gesture recognition, users can control their TVs by using simple hand gestures, such as swiping, pointing, and waving.

\subsection{Industrial Robots}

Gesture recognition technology is particularly useful in environments where workers need to keep their hands free, such as in cleanrooms, hazardous areas, or when wearing protective gear.

ABB's YuMi Robot is a collaborative robot that makes use of gesture recognition technology to improve its usability and functionality in industrial manufacturing environments. The robot's gesture-based programming allows operators to teach it new tasks simply by guiding its arms and hands through the required movements. 

\subsection{Virtual Reality}

The hand tracking feature of the Quest Oculus VR headset enables you to use your hands to navigate through various menu options. The motion data is used as input to decide what action to take.

UltraLeap motion is integrated with virtual and augmented reality to let users pick up objects without the use of a gadget.
The Leap Motion Controller is a compact USB device that lets users enter gesture commands into an application by using infrared LEDs to track hand and finger movements.

\section{Role of Gestures in Sign Language} 

Gestures play a crucial role in sign language as they are the basic units of communication in this visual language. Signers use gestures to create signs, which are combinations of movements, shapes, and locations that have specific meanings. For example, the sign for "happy" may involve a smiling facial expression, an upward movement of the hands, and a specific handshape.

Gestures are also important in sign language because they can convey grammatical information. Gestures can also be used to show tense, aspect, and other aspects of grammar.

Gestures allow signers to convey complex information and grammatical nuances, and they form the foundation of this unique visual language.

\section{Common challenges faced}

\begin{itemize}


\item Variability in human gestures: Variability in gestures due to various factors(culture, gender, age, lighting, ambiguity etc) makes it difficult to design a universal gesture recognition system that can accurately interpret different types of gestures.

\item Complex hand movements: Hand movements are complex and involve the use of multiple joints and muscles and hence requires a sophisticated algorithm that can analyze the subtle changes in hand position and movement for interpretation.

\item Limited training data: Training a gesture recognition system requires a large amount of data. However, there is a limited amount of training data available, which can make it difficult to create accurate models.

\item Real-time processing: Real-time processing is necessary for many applications of gesture recognition, such as gaming and virtual reality. However, real-time processing requires high-speed computing and processing power, which can be a challenge for some devices.


\end{itemize}

\section{Conclusion}
In conclusion, this survey paper has explored the concepts of gestures and gesture recognition systems, their applications and methodologies. It compares and contrasts various tools and techniques for perceiving gestures. It also analyses various data sources and corresponding data acquisition techniques. We have seen that gestures are a fundamental means of communication and can convey a wide range of meanings and emotions. Gesture recognition systems enable machines to interpret and respond to human gestures, which has numerous applications in various fields, including healthcare, gaming, and robotics.

In particular, gesture recognition systems have proven to be invaluable in sign language recognition and translation, facilitating communication between hearing and deaf communities. By recognizing and interpreting sign language gestures, these systems enable deaf individuals to communicate more effectively and bridge the communication gap between deaf and hearing communities.

Overall, this survey paper provides a comprehensive overview of the state-of-the-art in gesture recognition systems and highlights the current challenges and future directions in this field.

\end{document}